\documentclass{ecai}
\usepackage[linktocpage,colorlinks]{hyperref}			% PDF hyperlink
\usepackage{float}
\usepackage{amsmath}
\usepackage[ruled]{algorithm2e}
\usepackage{subcaption}
\usepackage{times}
\usepackage{graphicx}							% To include graphics
\usepackage{latexsym}
\newcounter{ntnucounter}% NTNU footnote counter
\newcounter{exposedcounter}% EXPOSED footnote counter

\begin{document}

\title{FishNet: A Unified Embedding for Salmon Recognition}

\author{Bjørn Magnus Mathisen \institute{Department of Computer Science, Faculty
    of Information Technology and Electrical Engineering, Norwegian University
    of Science and Technology
    email: bjornmm@ntnu.no}\setcounter{ntnucounter}{\value{footnote}}
  \institute{EXPOSED Aquaculture Research Centre,
    Department of Computer Science, Norwegian
    University of Science and Technology, Trondheim, Norway,
    https://exposedaquaculture.no/}
  \setcounter{exposedcounter}{\value{footnote}}
  \and Kerstin Bach \footnotemark[\value{ntnucounter}] \footnotemark[\value{exposedcounter}]
  \and Espen Meidell\footnotemark[\value{ntnucounter}] \\
  \and Håkon Måløy\footnotemark[\value{ntnucounter}] \footnotemark[\value{exposedcounter}]
  \and Edvard Schreiner Sjøblom\footnotemark[\value{ntnucounter}]}

\maketitle
\bibliographystyle{ecai}

\begin{abstract}
  Identifying individual salmon can be very beneficial for the aquaculture
  industry as it enables monitoring and analyzing fish behavior and welfare. For
  aquaculture researchers identifying individual salmon is imperative to their
  research. The current methods of individual salmon tagging and tracking rely
  on physical interaction with the fish. This process is inefficient and can
  cause physical harm and stress for the salmon. In this paper we propose
  FishNet, based on a deep learning technique that has been successfully used
  for identifying humans, to identify salmon.We create a dataset of labeled fish
  images and then test the performance of the FishNet architecture. Our
  experiments show that this architecture learns a useful representation based
  on images of salmon heads. Further, we show that good performance can be
  achieved with relatively small neural network models: FishNet achieves a false
  positive rate of \textbf{1\%}and a true positive rate of \textbf{96\%}.
\end{abstract}

\section{Introduction}
\label{cha:Introduction}

The Atlantic salmon farming industry in Norway has experienced a massive growth
in the past four decades. The industry has gone from producing 4.300 tonnes of
salmon in 1980, to almost 1.240.000 tonnes in 2017 \cite{2018Akvakultur}. In
2017, the total economical results from salmon production was calculated to be
over 61 billion Norwegian kroner (NOK) \cite{2018Akvakultur}. This makes salmon
farming one of the most profitable industries in Norway, and it is considered as
one of the most important industries in a post oil Norway \cite{nasjonalsjomat}.
However, the industry is still largely driven by manual labor. For example, the
total number of lice in a breeding cage is indicative of fish welfare and an
important metric for deciding whether delousing measures should be initiated.
Today's method for lice counting relies on manually inspecting individual fish
and then estimating the total number of lice in the cage from these numbers.
Other measurements such as disease and weight measurements also use similar
methods, based on a few individual fish measurements. These methods are highly reliant on
the fish inspected to be representative for the total population within the
cage. However, salmon is a schooling fish and organize themselves according to
hierarchical structures \cite{Assesing_swim, CUBITT2008529}. This means that
different types of individuals will be present at different layers of the
school. As the sampling methods used in the industry relies on small samples the
methods are prone to selecting the same type of individuals for inspection every time.
This could result in skewed estimations and lead to wrong operations being
performed. As these operations are often both costly and harmful for the fish,
large economic gains can be made from more precise estimates.

To improve measurement quality, a method of ensuring that different individuals
are measured every time is needed. Previous attempts at solving this problem
have included a variety of techniques. However, the techniques have almost
exclusively relied on physical engagement with the salmon. The techniques
include surgical implantation of tags and external mutilation, such as
fin-clipping, freeze brands, tattoos, visible implant tags, and external tag
identifiers attached by metal wire, plastic, or string \cite{merz}. This is a
problem both from an animal welfare and product quality perspective. Bacterial
growth and unpleasant sensory properties has shown to increase more quickly in
salmon experiencing stress in their lifetime prior to being slaughtered. This
results in reduced shelf life of the finished product \cite{Hansen2012}. A
computer vision method for uniquely identifying individuals would solve this
problem by minimizing the impacts from invasive techniques.

In this paper, we introduce an approach for accurately identifying individual
salmon from images, using a deep neural network called FishNet. By accurately
identifying individual salmon, we can ensure that no salmon is measured multiple
times, thereby guaranteeing a more accurate estimation of the total population.
Our approach is based on FaceNet \cite{schroff2015facenet} and DeepFace
\cite{taigman2014deepface} which have been proven to work well in the field of
face verification in humans. These networks are able to verify the identity of
people in images with human levels of accuracy. They have also been shown to be
robust to noise in terms of changing lighting conditions. By training a similar
architecture on images of fish rather than humans, we enable accurate identity
predictions without physical interaction.

Being able to track salmon at an individual level could enable tracking a single
individual throughout its lifespan, from salmon spawn to finished product,
linking salmon fillets to the life-story of the individual. Other opportunities
include monitoring individual weight development, treating salmon only when the
need arise and delousing only the individuals that suffer from lice, thereby
preventing unnecessary harm to healthy salmon. Individual salmon tracking could
also enable new research areas that require monitoring of individuals over time
such as feeding behavior, detection of diseases and social behavior. FishNet can
facilitate such research through offering a non-invasive and efficient approach
to identifying salmon.

The rest of this paper is structured in the following way. In Section
\ref{background:related} we outline the current state of the art within the
problem area of individual recognition of salmon. And as a results of the method
chosen to solve the problem we also outline the current state of the art of
using machine learning to identify individuals from pictures. Following this, we
present our approach to the problem of individually recognizing salmon in
Section \ref{sec:fishnet}. The dataset used for evaluation and the evaluation of
our proposed solution are presented in \ref{sec:ResearchAndResults}. We present
a discussion of our results in Section \ref{sec:discussion}. Finally
Section \ref{sec:conclusion} presents our conclusions and thoughts on future
directions of research for this work.

\section{Related Work}
\label{background:related}

Since the problem we address is inter-disciplinary, related work is two-fold: one
area of research covers the detection and identification of fish and salmon in
particular while the other one focuses on the classification of images. In this
section we’ll discuss the relevant work representing the state-of-the-art.

There has only been very limited work conducted to identify unique fish/salmon
without engaging directly with the fish. Earlier attempts of uniquely
identifying salmons have relied on insertion of RF-ID chips or other physical
marking systems \cite{sintefsalmid}. This is approach is only feasible in
research settings and should be minimized as it potentially injures the fish. In
real-world deployments with hundreds of cages and millions of fish a more
scalable approach is desirable. Throughout the recent years the field of
automatically identifying salmon has grown as the fish-farming industry
collaborate more and more with data-driven approaches. Especially in Norway,
projects such as the Exposed Aquaculture Operations Center for research based
innovation\footnote{\url{https://exposedaquaculture.no/en/}} or the Seafood
Innovation Cluster\footnote{\url{http://www.seafoodinnovation.no/}} emphasize on
applying Internet of Things, Big Data and Artificial Intelligence methods.

\begin{figure}[ht]
\centering
  \includegraphics[width=0.3\textwidth]{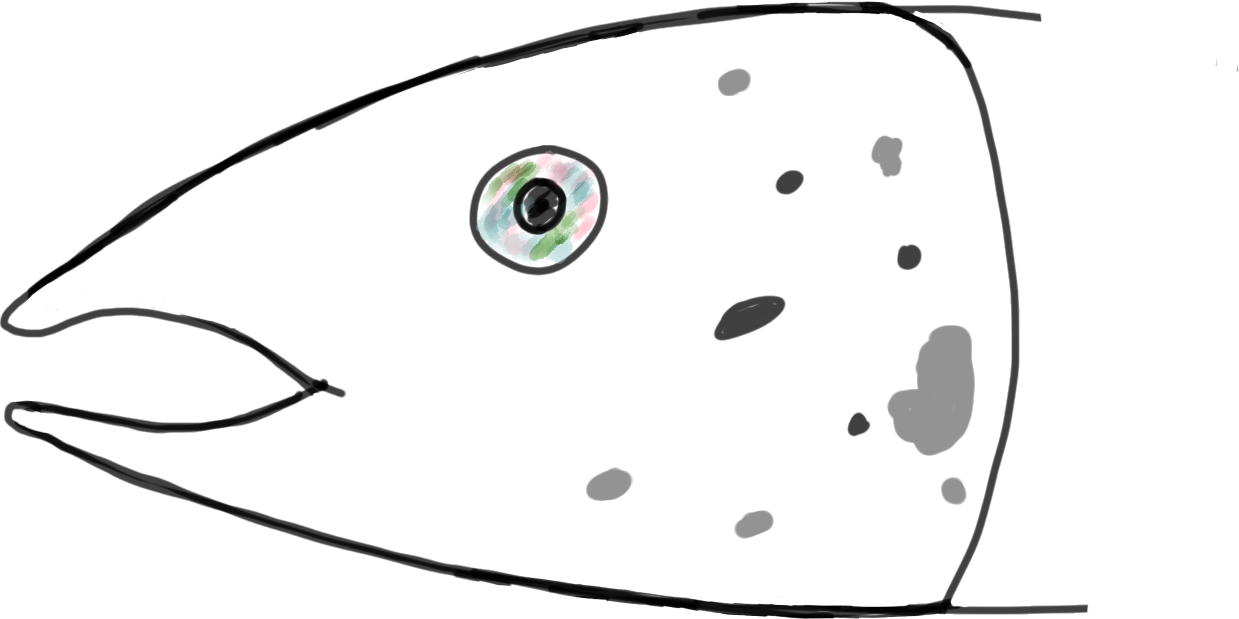}
  \caption{Melanin patterns on a salmon head.}
\label{fig:salmidmatch}
\end{figure}

SINTEF SalmID \cite{sintefsalmid} is a study that investigated the possibility
of recognizing individual salmon based on the assumption that each individual
has an unique pattern. They found that there was done little work on this area
regarding Atlantic salmon, but point at other work using the melanophore pattern
of different animals to uniquely identify them. The constellation of such
melanin patterns on the head of the fish can be utilized for identification. In the
SalmID approach the recognition part is based on manually selected features of
the salmon rather than learned representations.

Additional work utilizing melanophore patterns has been presented by Hammerset
\cite{ivarhammerset} who apply deep neural networks to discover the location of
salmon heads and eyes. In this work a simple blob detection algorithm is used to
discover the melanophore spots. The locations of the spots and the eye are then
translated into a polar representation which is saved in a database with the
identity of the salmon. On the test set with images from 333 individuals the
algorithm recognized 40.4\% (5922 of 14652 images) of the images as belonging to
an individual salmon, of these 40.4\% the algorithm correctly identified the
individual with an accuracy of 99.7\% (5902 of 5922). Thus the total test
accuracy was 40.2\% (5902 of the total 14652 images classified as the correct
individual)

Identifying individuals among humans has been an active research field for a
long time. Earlier work has been based on eigen value analysis of data matrices
such as EigenFaces \cite{eigenfaces} and its successors in FisherFaces
\cite{belhumeur1997eigenfaces} and Laplacianfaces \cite{he2005face}.

More recent woork on individual recognition is based on deep learning approaches
such as the model presented in the DeepFace paper \cite{taigman2014deepface} in
which they are making the images of faces more uniform (frontalization). These
frontalizations are fed into a convolutional layer followed by a max pooling
layer and another convolutional layer. According to the authors, these three
layers mainly extract low level features and make the network robust to local
translations. The last convolutional layer is followed by three locally
connected layers. This is done because the different regions of an aligned image
have different characteristics, so the spatial invariance assumption of
convolution does not hold. An example of this is that the eyes of a person will
always be above the nose. The final two layers of the network they use are fully
connected. These layers are able to identify relations between features in
different locations in the feature maps. The first fully connected layer is used
as the face representation vector, and the output of the second one is fed into
a softmax which produces a class distribution for an input image. To verify
whether two images are of the same person, the authors propose three approaches:
(1) an unsupervised method in which the similarity of two images is simply
defined as the inner product of the two representation vectors, (2) a weighted
$\chi^2$ distance in which the weight parameters are learned using a linear
support vector machine and (3) a siamese network, in which the network (except
the top layer used for softmax classification) is duplicated. One image is fed
into each part of the network and the absolute difference between the feature
vectors is computed. A fully connected layer is added and the network is trained
to predict a single logistic unit (whether the images are of the same person).
Training is only enabled for the new layers, and they are trained using standard
cross entropy loss. All three methods yielded good results compared to the
state-of-the-art at the time. The siamese network approach required a lot more
training data to avoid overfitting compared to the other approaches.

A related approach has been presented as FaceNet \cite{schroff2015facenet} which
describes a system that learns and optimizes a vector representation directly,
rather than extracting the representation from a bottleneck layer (like
DeepFace). FaceNet learns a 128-dimensional feature vector (embedding) that
represents a face. Unlike the DeepFace approach there is no 2D or 3D alignment
done on the images. FaceNet is a variant of a Siamese Neural Network (SNN)
originally proposed by Bromley \cite{bromley1994signature}. In contrast with the
original SNNs FaceNet uses triplet loss to train the network. The network is
presented with three images (the anchor image, the positive image (same person,
but different image and the negative image (image of any other person).
According to the authors, it is important to select triplets that are hard for
the model to correctly discriminate, to ensure that the network converges as
quickly as possible during training. The triplets are chosen from within each
mini-batch, and all anchor-positive pairs are used in combination with negative
examples. The authors describe several different deep neural network
architectures, where the major differences between them are the number of
trainable parameters. The number of parameters in the networks range from about
4 million to 140 million. When evaluating the networks the $L_2$-distance
between two images is compared. If the distance is above a certain threshold
they are classified as different. According to the authors they are able to
reduce the error reported by the DeepFace paper by a factor of seven. The
smaller inception networks perform nearly as good as the very deep networks.

\section{The FishNet Approach}
\label{sec:fishnet}
\begin{figure}[H]
  \centerline{\includegraphics[width=\columnwidth]{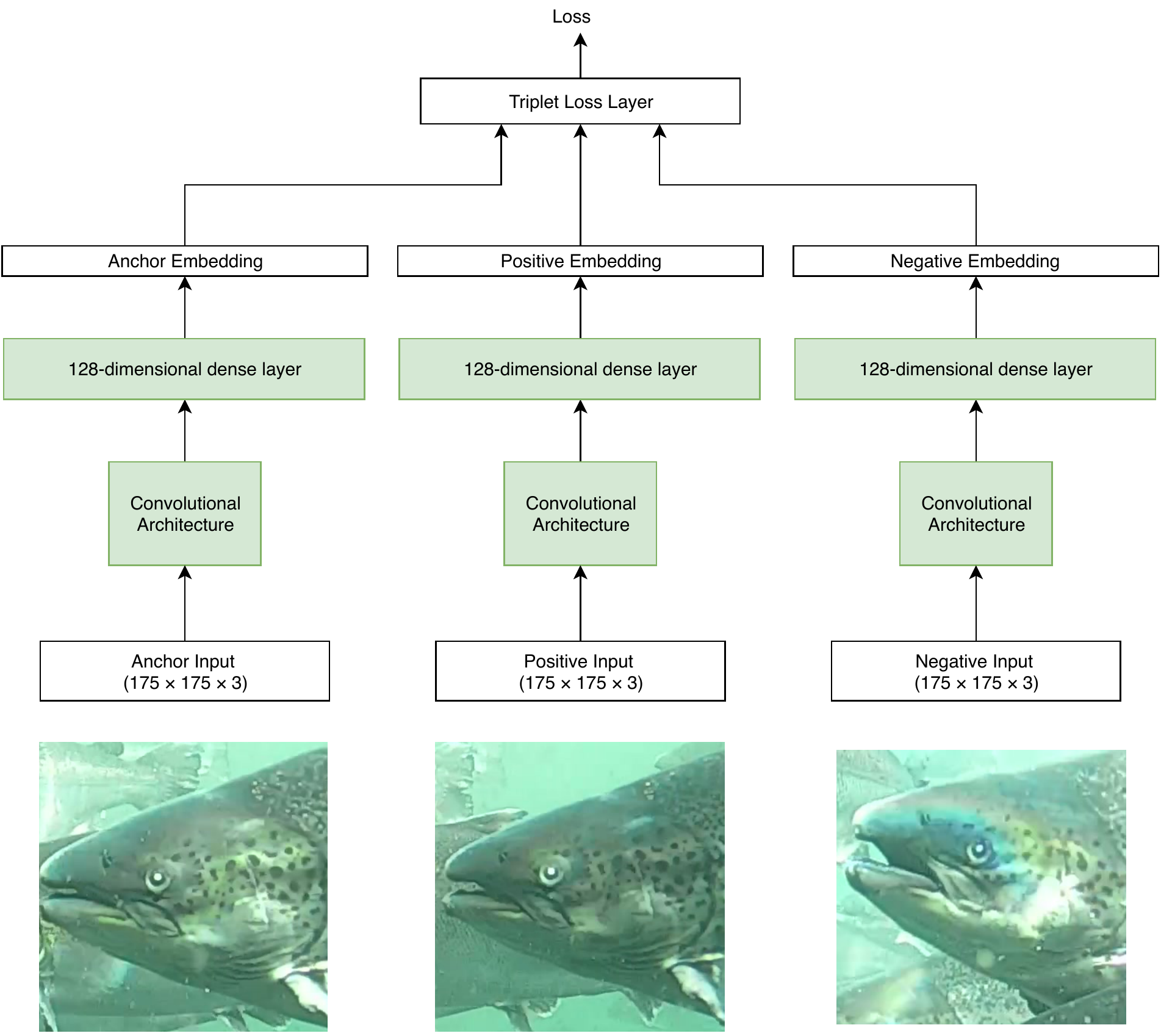}}
  \caption[Generic Architecture with Triplet Loss]{Generic architecture with
    triplet loss. Parts of the network with shared weights are colored green.
    the input size is the size of the images (175x175x3) and the output is the
    128-length embedding vector. The differences between the model architectures
    tried in our experiments is how the convolutional architecture is modeled,
    and the size of that convolutional model. This figure shows an example of salmon
    heads, with the anchor input to the left, positive input (same individual
    as the anchor input) in the center and finally the negative input
    (different indivdual than the anchor input).}
  \label{fig:genericarchitecturetriplet}
\end{figure}

To recognize individual salmons we adapt the FaceNet \cite{schroff2015facenet}
architecture and training method. FaceNet is a type of Siamese neural
network\cite{bromley1994signature,Mathisen2019} which has two datapoints as
input, and the output is the distance between them. This can also be extended
to work on e.g. triplets of data points, outputting more than one distance.
FaceNet is trained on a dataset consisting of triplets consisting of a anchor
data point, a positive data point and a negative data point. The anchor data
point with a given label, the positive data point is a different data point with
the same label, in contrast the negative data point has a different label.
Figure \ref{fig:genericarchitecturetriplet} illustrates this with three example
images of salmon, two of which are from the same individual salmon, while the
third image is of another individual salmon.

\begin{figure}[H]
  \begin{center}
    \includegraphics[width=\columnwidth]{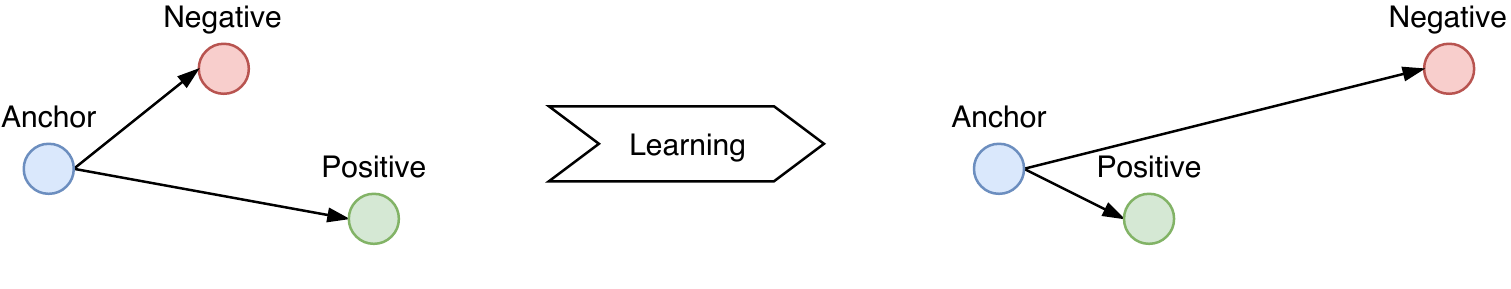}
    \caption[Triplet Loss]{Triplet loss minimizes the distance between images of the
      same salmon and maximizes the distance to images of other salmon (adapted
      from \cite{schroff2015facenet}).}
    \label{fig:tripletloss}
  \end{center}
\end{figure}

The goal of training FaceNet is to minimize the distance between the anchor and
positive data point, while maximizing the distance to the negative data point.
This training process is illustrated in Figure \ref{fig:tripletloss}.

To compute the loss during the training, a custom triplet loss layer was used.
Equation \ref{eq:tripletlosslayer} defines how the loss $L$ is computed for a
minibatch of size $m$.

\begin{equation} \label{eq:tripletlosslayer}
\begin{aligned}
  L &=
  \sum_{i}^{m}\left[\left\|\hat{x}_{i}^{a}-\hat{x}_{i}^{p}\right\|_{2}^{2}-\left\|\hat{x}_{i}^{a}-\hat{x}_{i}^{n}\right\|_{2}^{2}+\alpha\right]_{+}
\end{aligned}
\end{equation}
Here $\hat{x}=f(x)$ is the embedding of image $x$, $x^{a}$ is the anchor data
point, $x^{p}$ is the positive data point, $x^{n}$ is the negative datapoint and
$\alpha$ is a parameter that encourages better learning.

This is identical to how the triplet loss is defined in the
FaceNet \cite{schroff2015facenet}. The loss computes the distance between the
anchor and the positive, and the anchor and the negative. The goal is to have
the positive distance be smaller than the negative distance. The difference
between the positive and negative distance are summed. To encourage larger
distances the margin $\alpha$ is added to the loss function. To avoid negative
loss, the loss is set to the maximum of the loss of the triplet and 0.

Careful triplet selection is important \cite{schroff2015facenet} for the training
process of the network. The training goal of the algorithm is to ensure that the
embeddings of two images (anchor and positive) of the same salmon are closer to
each other than any images of other salmon (negatives) by a margin $\alpha$. In
our experiments, the value for $\alpha$ was set as $0.2$, the same
as used in the FaceNet paper.

To ensure effective training, it is important to select triplets that violate
this constraint. To do this, the method computes the embeddings for images
during training, and then select samples only among the triplet that violate
this training samples. For efficiency purposes, this is done within each batch.
First, a random set of salmon images are sampled from the training dataset. Then
the images are fed through the network to generate embeddings. Finally, the
embeddings are used to select triplets where the difference between the negative
and positive embeddings are within $\alpha$. Algorithm \ref{algo:tripletsel}
describes this process. \newline

\begin{algorithm}[ht]
  \SetAlgoLined
  \KwIn{embedding vectors}
  \KwIn{number of fish}
  \KwIn{number of embeddings per fish}
  \KwIn{$\alpha$}
  \KwData{triplets = []}
  \ForEach(){fish} {
    \For{anchor in embeddings of current fish}{
      negative distances = $L_2$-distances from anchor to embeddings of other fish\\
      \For{positive in remaining embeddings of current fish}{
        compute distance between anchor and positive\\
        negatives = find all negative embeddings where $negative\_dist - positive\_dist < \alpha$\\
        select a random negative from negatives and append (anchor, positive, negative) to triplets
      }

    }
  }
  shuffle triplets\\
  return list of triplets
  \caption{Triplet selection}
  \label{algo:tripletsel}
\end{algorithm}

Using Algorithm \ref{algo:tripletsel} to select the triplet used for training,
we ensure that training is performed on triplets the network can learn from.
Using triplets that already satisfy the constraint of $\alpha$ would not
contribute to further training, and only slow down the process. Calculating the
hardest triplets for the entire dataset every epoch would be computationally
very slow. Additionally, if we were to select the hardest triplets every time it
could cause poor training. This is because selection of hardest triplets would
be dominated by for example mislabeled or low quality images.

\subsection{Neural Network Architectures}
\label{sec:nnarchitectures}

During our experiments, we trained different neural network architectures to
produce embeddings. All the networks shared a general architecture of a
convolutional neural network where the top layer (classification layer) was
replaced by a 128-dimensional dense layer to represent the embedding of the
input image. Figure \ref{fig:genericarchitecturetriplet} shows an illustration
of this architecture, which is used to compute the embedding for one image.
Table \ref{tbl:archsize} we show the different types of architectures we
evaluated as part of this work. This was done to investigate the effect of using
different convolutional architectures and model sizes to produce embeddings. The
corresponding results to the architectures listed in this table is listed in
Table \ref{tbl:model_auc_tpr}.

\begin{table}
\centering
\begin{tabular}{lllll}
Network Architecture           & \# Parameters & Pretrained with \\ \hline
FishNet1 (Inception ResNet v2) & 55M           & ImageNet  \\
FishNet2 (MobileNet v2)        & 2.4M          & ImageNet  \\
FishNet3 (VGG-16)              & 15M           & ImageNet
\end{tabular}
\caption[Neural network architectures]{The different neural network architecture
  models used in the experiments. From a large model (FishNet1 based on
  Inception ResNet v2) to a model that is 20 times smaller (FishNet2 based on
  MobileNet v2) that can be deployed on a mobile device.}
\label{tbl:archsize}
\end{table}

To train the network using triplet loss, the network needs to use more than one
image at once. To achieve this, the convolutional and embedding parts need to be
replicated once for each image. Note that the weights are shared between the
instances. The output from the embedding layers is fed into a custom layer that
computes the triplet loss, which in turn is used to train the model. Figure
\ref{fig:genericarchitecturetriplet} illustrates the model used for training.
Table \ref{tbl:archsize} shows the neural network architectures used in the
experiments.

All models were initialized with the convolutional weights pretrained on the
ImageNet dataset \cite{imagenet_cvpr09}. The assumption being that features
learned for image classification may be a useful starting point for learning how
to distinguish salmon from each other, thereby reducing the amount of training
data needed to train the models.

\section{Dataset and Evaluation}
\label{sec:ResearchAndResults}
As far as we know, there is currently no data set of labeled fish to use for
training and evaluating methods for identity recognition. Thus, to evaluate the
FishNet method we needed to create a dataset of labeled pictures of salmon
heads. To do this we aquired a video clip of salmon swimming from
SeaLab\footnote{\url{https://www.sealab.no/}}.

\begin{figure}[ht]
  \centering
  \begin{subfigure}{0.7\columnwidth}
    \includegraphics[width=0.9\linewidth]{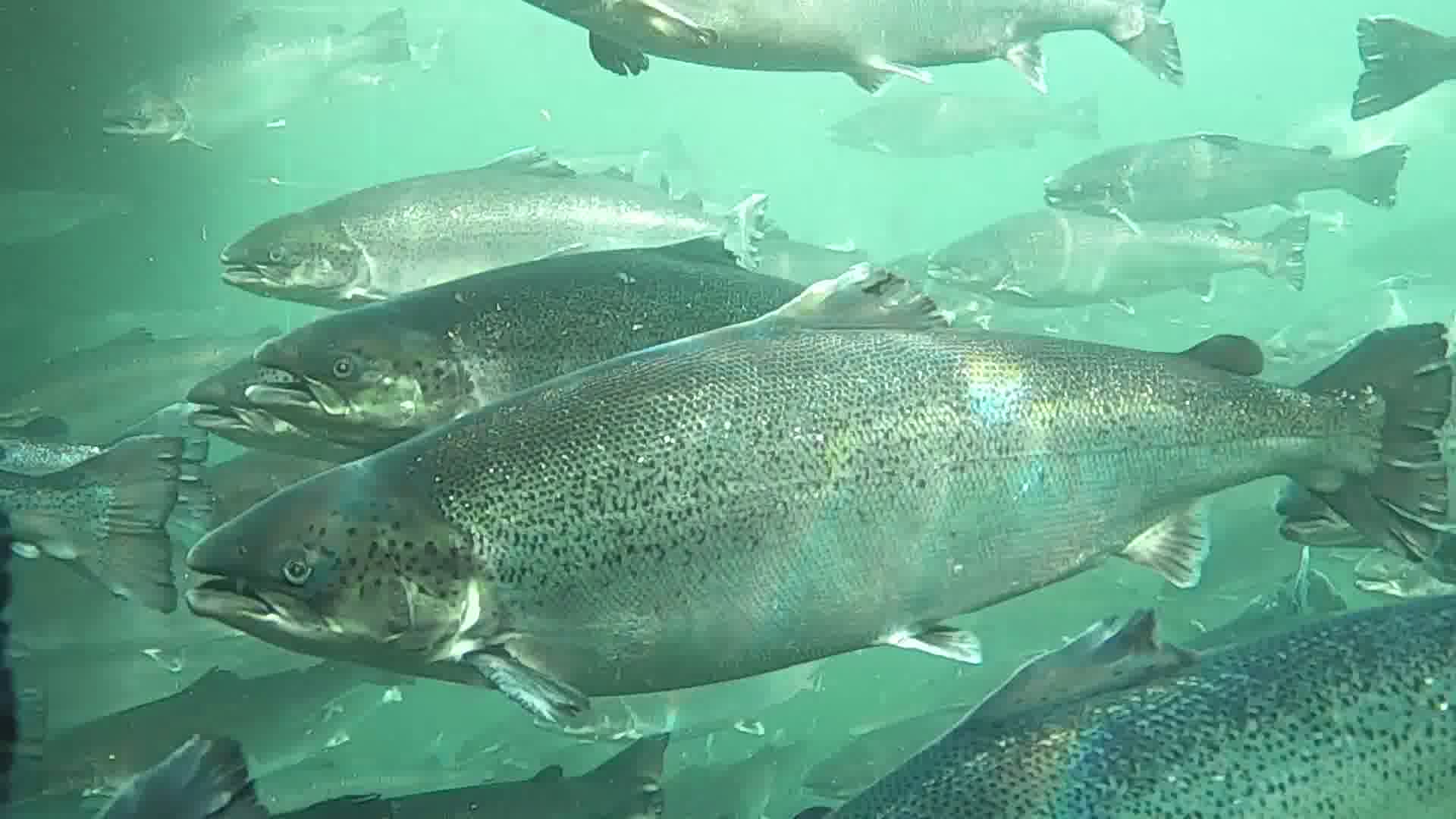}
    \caption{An example of a frame from the original video.}
    \label{fig:originalframe}
  \end{subfigure}
\qquad
  \begin{subfigure}{0.7\columnwidth}
    \includegraphics[width=0.9\linewidth]{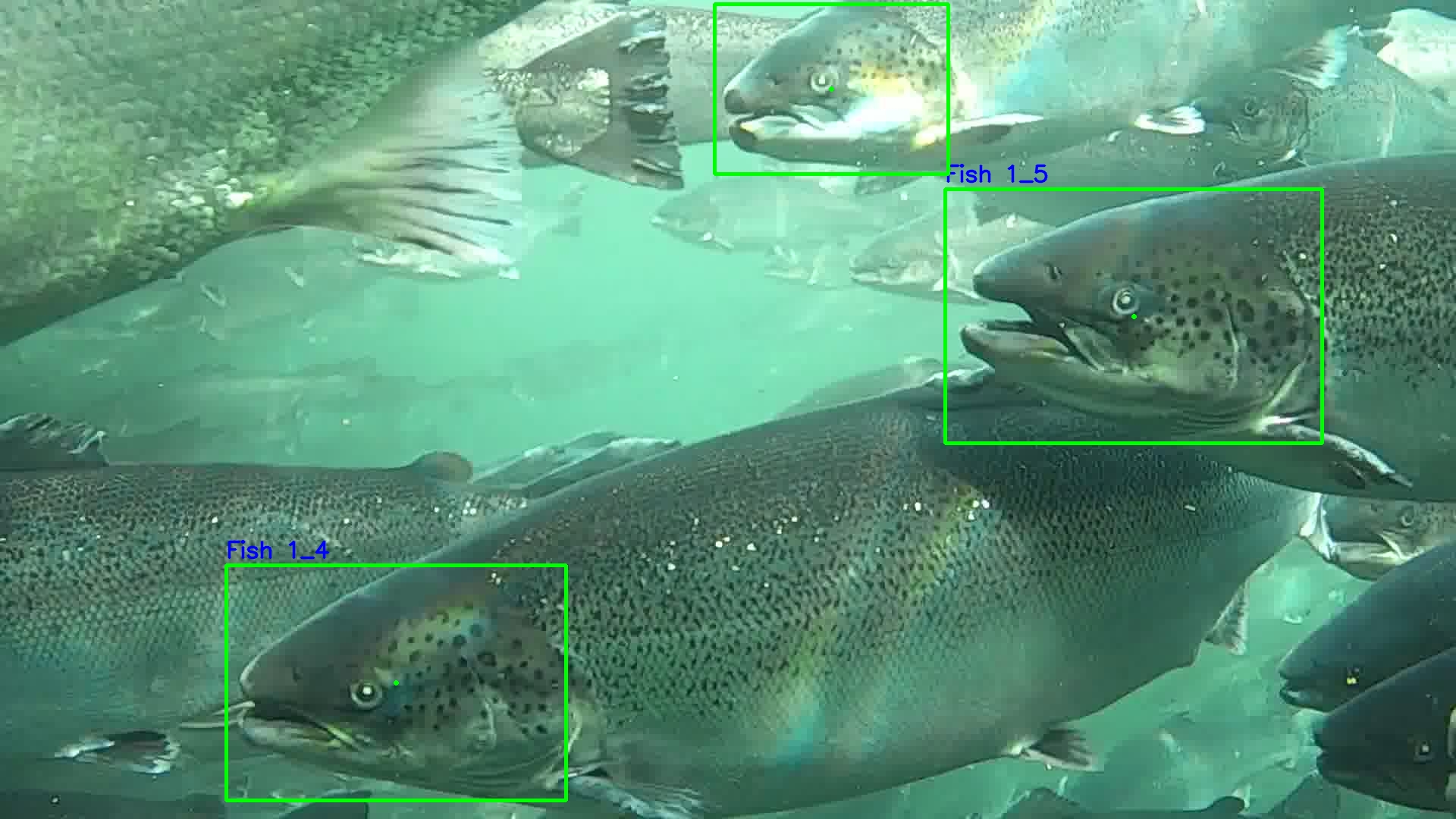}
    \caption{An example of bounding-boxes of salmon heads detected by the YOLOv3
      model trained to detect salmon heads.}
    \label{fig:boundingboxes}
  \end{subfigure}
  \caption{Overview of different stages the dataset creation.}
  \label{fig:datasetoverview}
\end{figure}

\begin{figure}[hb]
  \begin{center}
    \includegraphics[width=0.9\columnwidth]{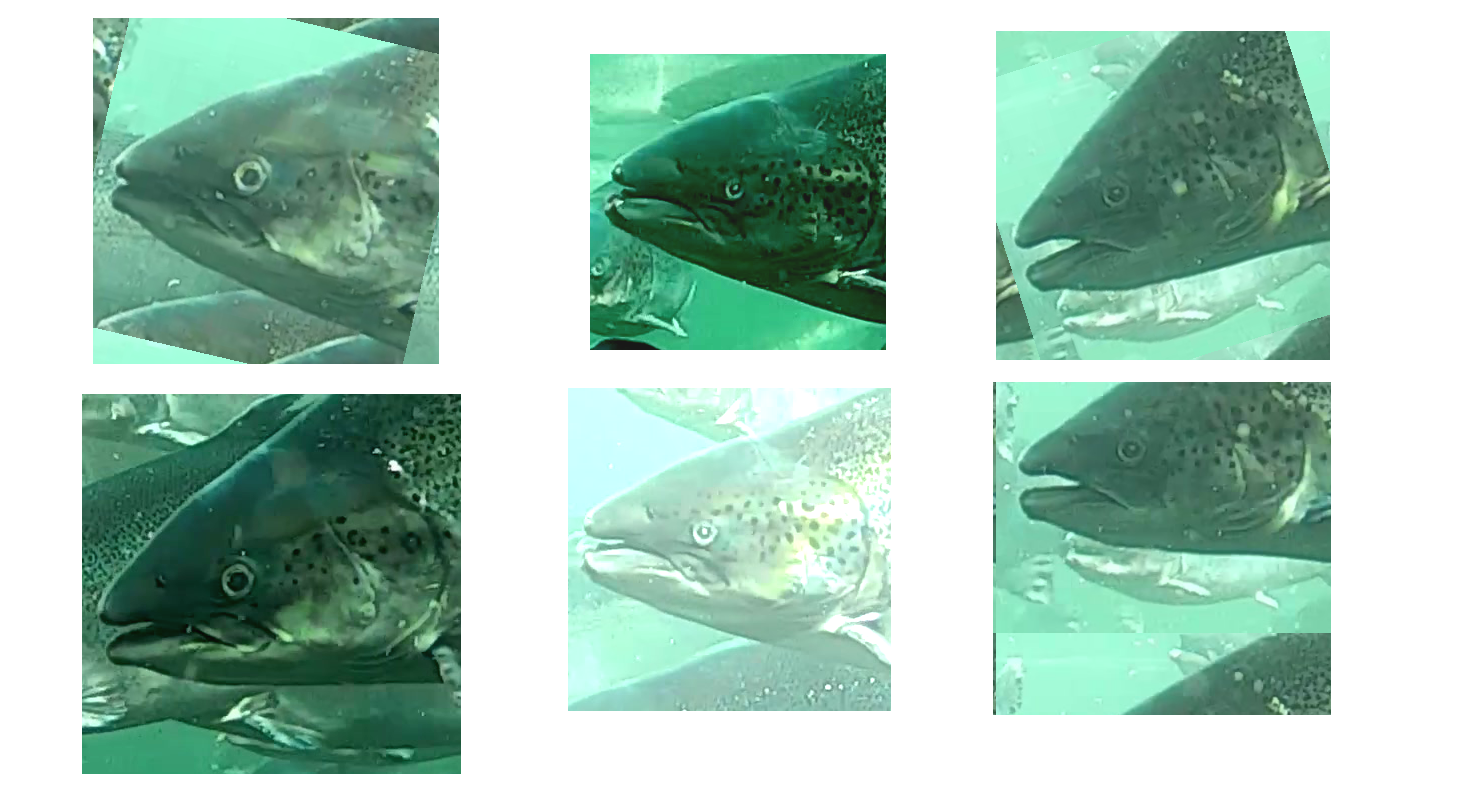}
    \caption{Augmenting images during the dataset creation. The top right and
      top left images show tilting augmentation. Centre bottom is color shifting
    (making the image brighter).}
    \label{fig:augmentation}
  \end{center}
\end{figure}

The original data was a video stream of salmon swimming across the view of the
camera. The video was filmed at 30 FPS (frames per second) meaning we had 30
images per second of video. Figure \ref{fig:originalframe} shows a frame
captured from this video. Salmon heads in the images were marked manually with a
bounding-box tool. After manually labeling approximately 500 bounding-boxes as
salmon heads the bounding-boxes were used to train a YOLOv3 (\cite{yolov3})
network to recognize salmon heads. This YOLOv3 model was then used to create
bounding-boxes on every salmon head in all video frames, as seen in Figure
\ref{fig:boundingboxes}. Figure \ref{fig:genericarchitecturetriplet} and Figure
\ref{fig:fishmatrix} shows examples of the resulting cropped bounding-boxes of
two salmon heads. These bounding boxes is then extracted as a $175x175$ image.
These images are then clustered to achieve clusters of images for each
individual salmon. Equation \ref{eq:dbscandistance} describes the distance
function used in the clustering algorithm. If two bounding-boxes are in the same
frame, the distance is set to an arbitrarily high value. If the bounding-boxes
are not in the same frame, the intersection over union is measured to check how
closely the bounding-boxes overlap. Then a temporal distance is added by
computing a weighted distance of the frame numbers. This is done to ensure that
overlapping bounding-boxes in frames next to each other receive a low distance
value. These distance metrics are combined into one single distance (Equation
\ref{eq:dbscandistance}) metric which is used by DBSCAN\cite{dbscan} to cluster
the images. This process produces clusters of images of the same individual
salmon. This approach works fairly well except in cases where a salmon
disappears behind a different salmon and then reappears again. In those cases it
is frequently misidentified as a new salmon. This problem was solved by manually
reviewing the labels, and replacing the labels for misidentified salmon.
% \vfill\eject

\begin{equation} \label{eq:dbscandistance}
  \begin{aligned}
    D(b1,b2) &= \begin{cases}
      \infty & : \delta_{f} = 0\\
      \frac{1-IOU(b1, b2) + }{2} & : \text{otherwise} \end{cases} \\
  \end{aligned}
\end{equation}
Here $\delta_{f} = b1_{frame} - b2_{frame}$ and
\begin{equation*} \label{eq:dbscandistancenext}
  \begin{aligned}
    \\\\
    IOU(b1, b2) &= \frac{\text{Intersection Area}}{\text{Union Area}}
  \end{aligned}
\end{equation*}

\begin{figure}[ht]
  \begin{center}
    \includegraphics[width=0.7\columnwidth]{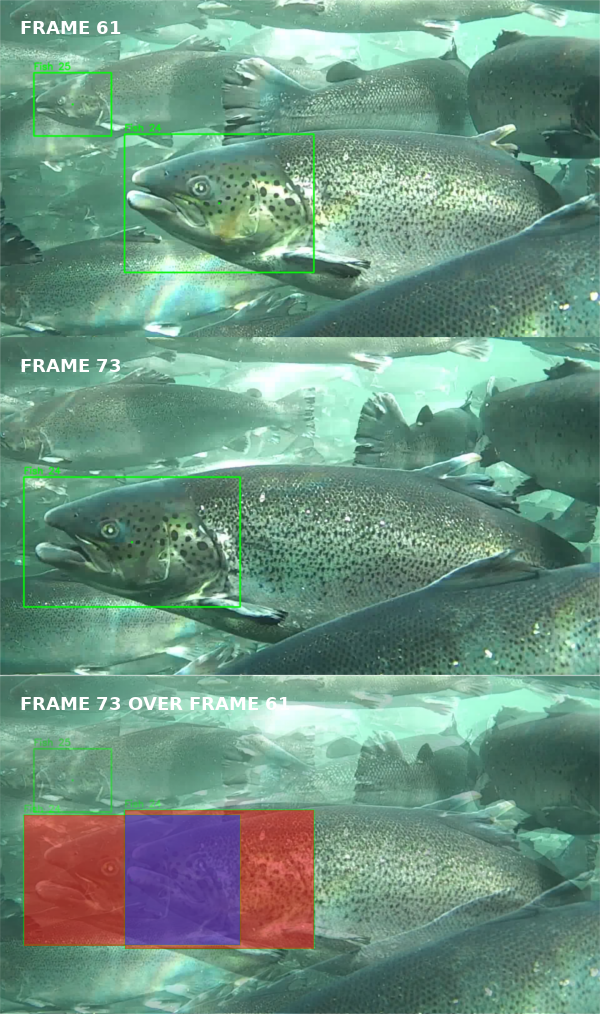}
    \caption[Illustration of IOU]{Illustration of IOU. The top image is frame 61
      in the video, the middle is from frame 73, and the bottom image shows the
      two images stacked on top of each other. Despite being 12 frames apart,
      the IOU is still quite high. The red and blue area is the union between
      the bounding boxes, and the blue area alone is the intersection.}
    \label{fig:iou}
  \end{center}
\end{figure}

After the salmon heads are extracted they were clustered and finally labeled.
This resulted in a dataset of 15000 images of 715 individual salmon. The images
were then augmented by tilting the image, moving the image vertically and
shifting the brightness of the image. Examples of these augmentations can be
seen in Figure \ref{fig:augmentation}. Five augmented images were created for
each original image, resulting in a data set containing a total of 225 000
images divided over 715 individuals. The data set was then divided into test and
training sets, with 90\% of the images being used for training and the remaining
10\% being used for testing.

\begin{figure}[ht]
  \centerline{\includegraphics[width=0.9\columnwidth]{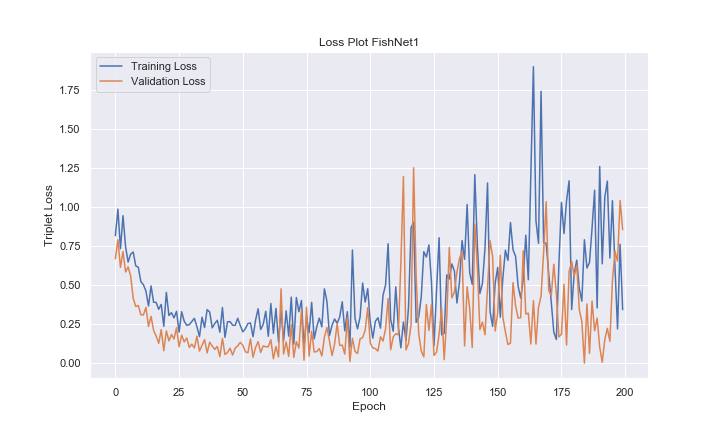}}
  \caption[The loss curves for FishNet1.]{The loss curves during training for FishNet1.}
  \label{fig:lossfishnet1}
\end{figure}

\subsection{Evaluation}
\label{sec:results}
\begin{figure}[hb]
  \centerline{\includegraphics[width=1\columnwidth]{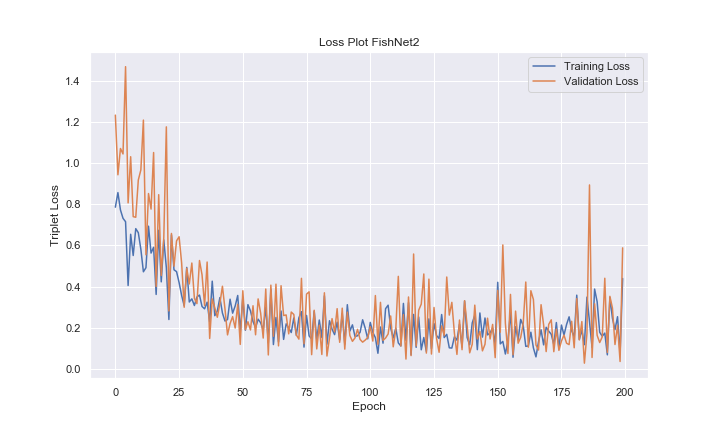}}
  \caption[The loss curves for FishNet2.]{The loss curves during training for FishNet2.}
  \label{fig:lossfishnet2}
\end{figure}

\begin{figure}[H]
  \centerline{\includegraphics[width=1\columnwidth]{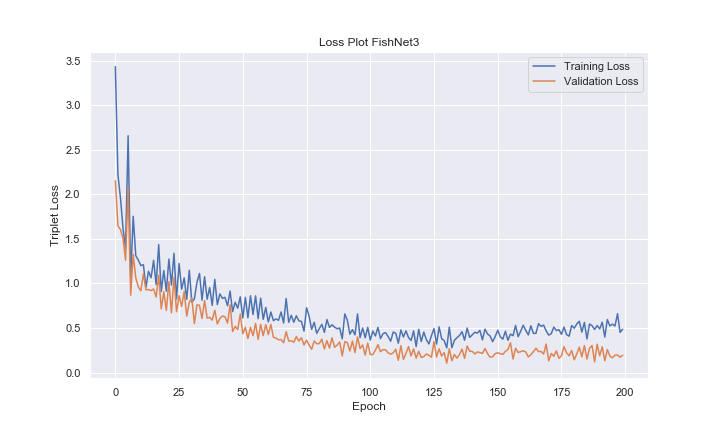}}
  \caption[The loss curves for FishNet3.]{The loss curves during training for FishNet3.}
  \label{fig:lossfishnet3}
\end{figure}

\begin{figure}[ht]
  \centerline{\includegraphics[width=1\columnwidth]{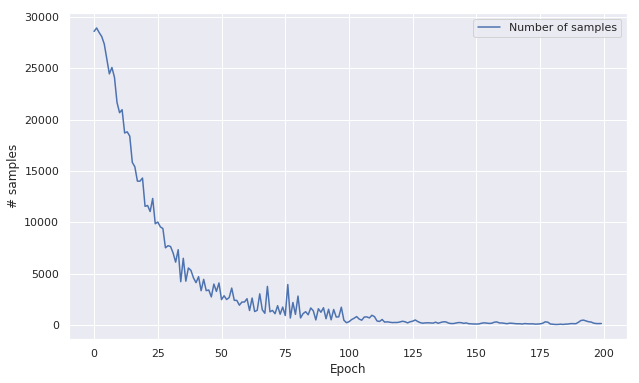}}
\caption[Number of triplets per epoch]{Number of triplets available for training
  each epoch for FishNet1. Towards the end of the training only about 100
  samples were available for training.}
\label{fig:nsamples}
\end{figure}

The experimental setup consisted of a single computer containing an AMD Ryzen
Threadripper 2920x 12-core CPU, two GeForce RTX 2080Ti GPUs and 128 GB of RAM.
The models were implemented using Tensorflow \cite{tensorflow2015-whitepaper}.
Figures \ref{fig:lossfishnet1}, \ref{fig:lossfishnet2}, and
\ref{fig:lossfishnet3} show the loss curves during the training of the three
models presented in section \ref{sec:nnarchitectures}. One notable observation
in the loss curves for FishNet1 is that both the training and validation loss
start to fluctuate and increase greatly towards the middle and end of training.
This occurs due to the nature of the triplet selection algorithm used during the
training phase. The algorithms only uses triplets that fail the triplet
constraint test described in section \ref{sec:fishnet}. This means that
if the model learns to separate salmon well, there are fewer triplets available
for training as the training progresses. By examining the training logs we can
see that this in fact happens. Figure \ref{fig:nsamples} shows show many of the
sampled triplets the network was able to use for training.

As seen in Figure \ref{fig:lossfishnet1}, when the models become increasingly
good at recognizing individuals the loss starts to fluctuate. This is most
likely due to the fact that the training process is down to a very small set of
triplets that is very hard to discriminate. When the training process seeks for
a model that can discriminate these last triplets, the loss value of the rest of
the dataset increases.

\begin{figure}[hb]
  \centerline{\includegraphics[width=1\columnwidth]{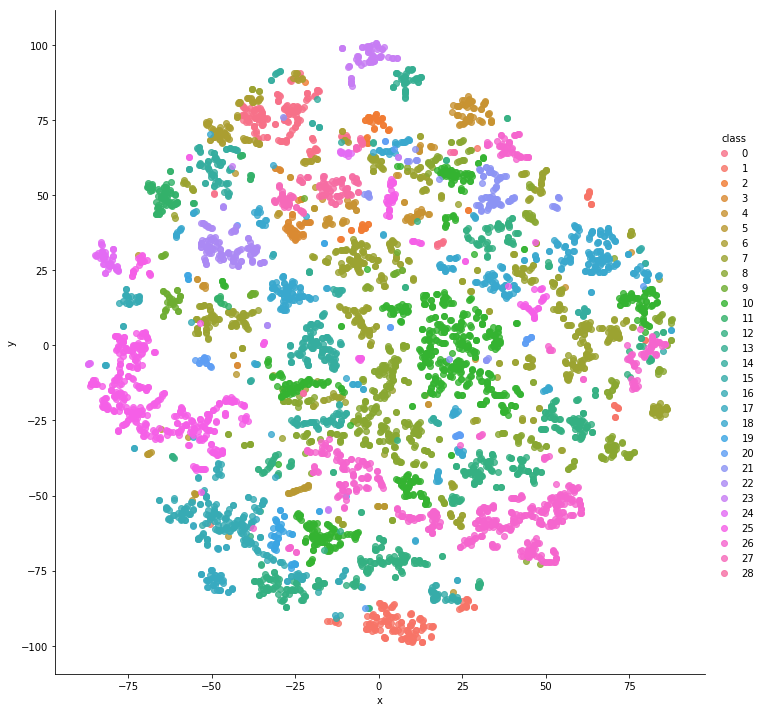}}
  \caption{T-SNE clustering of embeddings produced by a untrained FishNet1
    model. The slight clustering visible in the figure is an effect of the
    inherent clustering done by T-SNE.}
  \label{fig:tsnebefore}
\end{figure}

\begin{figure}[hb]
  \centerline{\includegraphics[width=1\columnwidth]{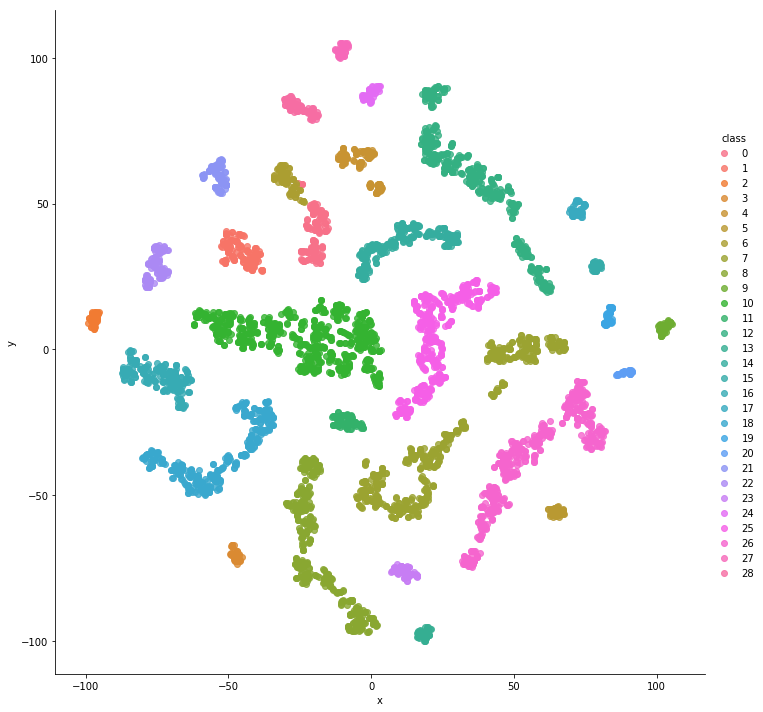}}
  \caption{T-SNE clustering of embeddings produced by a trained FishNet1 model.
    This is clearly a better clustering than shown in Figure
    \ref{fig:tsnebefore}, illustrating that the embedding process extracts
    useful signals for identifying individuals. }
  \label{fig:tsneafter}
\end{figure}

The goal of the face verification task is to easily be able to separate the
embeddings generated by different identities in the euclidean space. Figure
\ref{fig:tsnebefore} and Figure \ref{fig:tsneafter} illustrates how the
embeddings are distributed in the space before and after training. The points in
the plots are of $6000$ images from $29$ different salmon from the test set. The
models used are FishNet1 before and after $200$ epochs of training. As we can
see from the t-SNE-reduced plots the grouping of embeddings from salmon of the
same identity is far better after training. This indicates that the model is
able to learn some mapping from the images to embeddings.

To compute metrics such as true positive rate, false positive rate, accuracy
etc., a similarity threshold needs to be set. To compare the models we can
examine what the true positive rate (the sensitivity) of the system is at a set
false positive rate. We have compared the models where the false positive rate
is 0.01, that is, where 1\% of the negative samples are misclassified as
positive. As we can see in Table \ref{tbl:model_auc_tpr} FishNet1 and FishNet2
perform approximately equally with a true positive rate of about 96\%. FishNet3
performs significantly worse with a true positive rate of 87\%.

\begin{table}
\centering
\begin{tabular}{lll}
Network Architecture           & AUC    & TPR @ FPR = 10e-3  \\ \hline
FishNet1 (Inception ResNet v2) & 0.9977 & 0.964              \\
FishNet2 (MobileNet v2)        & 0.9974 & 0.961              \\
FishNet3 (VGG-16)              & 0.9919 & 0.870
\end{tabular}
\caption[AUC and TPR of models]{The area under the curve and true positive rate
  (measured when the false positive rate is 10e-3) of the models.}
\label{tbl:model_auc_tpr}
\end{table}

\begin{figure}[hb]
  \begin{center}
    \includegraphics[width=1\columnwidth]{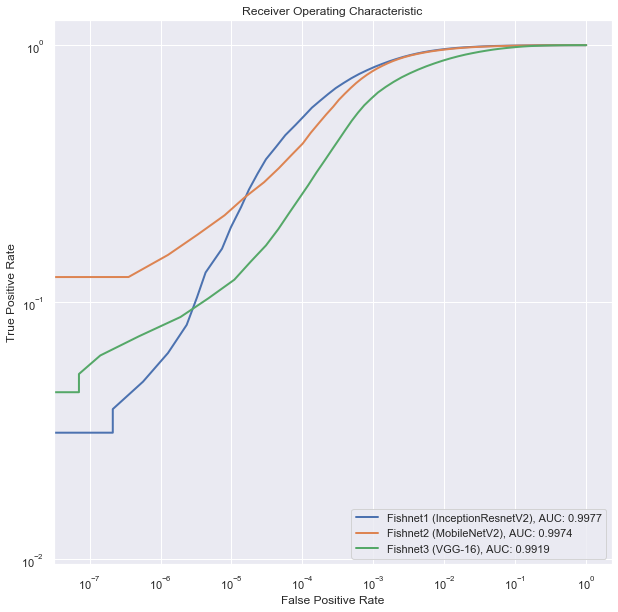}
    \caption[ROC Curve]{The ROC curves of FishNet1 (blue), FishNet2 (orange),
      and FishNet3 (green). The true positive rate and false positive rate is
      computed across similarity thresholds in the range [0.0, 2.0] in
      increments of 0.2. The model with the largest area under the curve has the
      best overall performance (FishNet1, with InceptionResnetV2). Note that the
      axes in the plot are in logarithmic scale.}
    \label{fig:roccurve}
  \end{center}
\end{figure}

Figure \ref{fig:roccurve} shows the ROC curve for the three models we tested. By
comparing the area under the curve we can compare the performance of the models
across all thresholds. As we can see FishNet1 and FishNet2 perform better than
FishNet3, with FishNet1 being the best of the models tested in our experiments.
It is interesting to note that the improved results of FishNet1 come at quite a
high computational cost compared with FishNet2, a network designed to be able to
run on mobile devices.

\begin{figure}[H]
  \begin{center}
    \includegraphics[width=1\columnwidth]{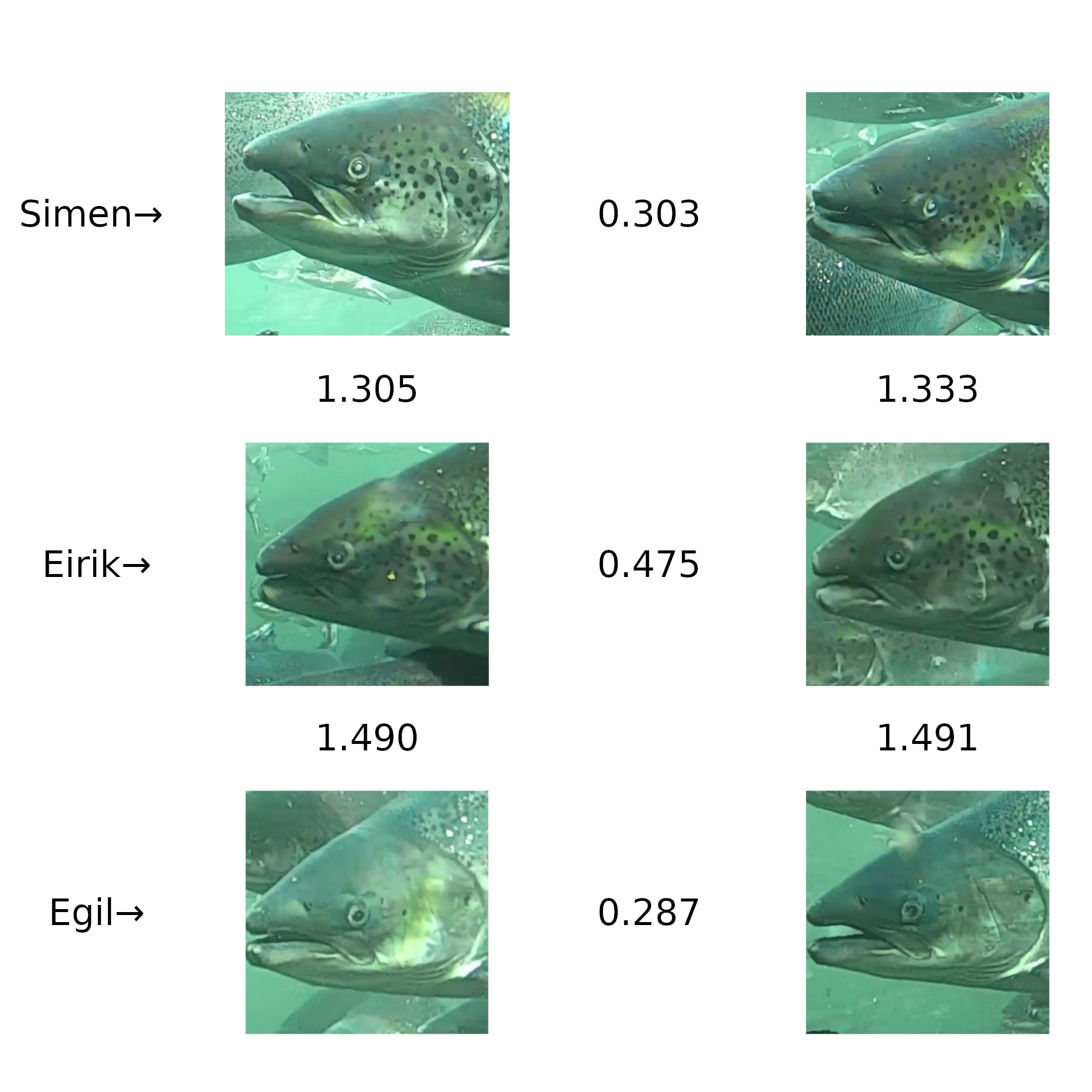}
    \caption[Fish distance matrix]{An illustration of the distances between six
      images from salmon with three different identities. Each row contains two
      images of the same salmon: ``Simen'' at the top, ``Eirik'' in the middle
      and ``Egil'' at the bottom. The average distance between the same salmon
      is $0.36$ while comparisons between different salmon average at $1.40$.}
    \label{fig:fishmatrix}
  \end{center}
\end{figure}

Lastly Figure \ref{fig:fishmatrix} shows an example of a visual evaluation of 2
images of 3 different salmon individuals (``Simen'', ``Eirik'' and ``Egil'').
This shows us that the calculated distances between different individuals are at
least three (on average $3.88$) times bigger than the distances between two
images of the same individual in this example.

\section{Discussion}
\label{sec:discussion}

The results shown in the Section \ref{sec:ResearchAndResults} demonstrate that
machine learning methods successfully applied for identifying humans from
pictures can also be used to identify individual salmon. However it should be
noted that this was done with frames extracted from a single video captured over
a short period of time. Thus the different frames representing the same
individuals in the data set created for this work are very similar. This is
somewhat amended by the augmentation done to the frames as described in Section
\ref{sec:ResearchAndResults}. However, the results from evaluating the method on
a data set with these augmentations does not enable us to conclude that the
method works under all conditions, or over longer periods of time. The video
used in this work is a video with very favorable conditions, both in terms of
light and water clarity. Training FishNet in more challenging conditions might
reduce the performance of the architecture. Thus adversarial regularization
using both artificial noise and adversarial examples could be beneficial or even
necessary for the architecture to handle such conditions, as this has been shown
to increase robustness of deep architectures \cite{adversarial_regularization}.
The fish may also be damaged mechanically or contract diseases which changes the
way individual fish looks over time. This could drastically affect the
performance of FishNet on such individuals.

\section{Conclusion}
\label{sec:conclusion}
In this paper we presented FishNet, a novel approach for individual fish
recognition using a convolutional deep neural network as part of a Siamese
neural network architecture based on FaceNet \cite{schroff2015facenet}. We
trained this model using images of salmon to make the model identify individual
salmon. FishNet achieves a false positive rate of 1\% and a true positive rate
of 96\%.

As future work we would like to investigate the model's ability to recognize
individuals from spawn to grown fish. We would also like to test if we can
increase performance by employing other variants of Siamese neural networks such
as eSNN\cite{Mathisen2019}. Finally, we would like to investigate what the
architecture is actually looking at when recognizing individuals.

\section{Acknowledgements}
This work is an extension of the MSc Thesis ``FishNet: A Unified Embedding for
Salmon Recognition''\footnote{\url{http://hdl.handle.net/11250/2628800}} by
Espen Meidell and Edvard Schreiner Sjøblom. This research has been funded by the
Research Council Norway, EXPOSED Aquaculture Research Center (grant number
237790) and the Norwegian Open AI Lab. In addition the data that formed the
basis for the data set was provided by Sealab
AS\footnote{\url{https://www.sealab.no}}

\bibliography{bibliography}

\end{document}